# Object Detection and Tracking Algorithms for Vehicle Counting: A Comparative Analysis

Vishal Mandal and Yaw Adu-Gyamfi

*Abstract*— The rapid advancement in the field of deep learning and high performance computing has highly augmented the scope of video-based vehicle counting system. In this paper, the authors deploy several state-of-the-art object detection and tracking algorithms to detect and track different classes of vehicles in their regions of interest (ROI). The goal of correctly detecting and tracking vehicles' in their ROI is to obtain an accurate vehicle count. Multiple combinations of object detection models coupled with different tracking systems are applied to access the best vehicle counting framework. The models' addresses challenges associated to different weather conditions, occlusion and low-light settings and efficiently extracts vehicle information and trajectories through its computationally rich training and feedback cycles. The automatic vehicle counts resulting from all the model combinations are validated and compared against the manually counted ground truths of over 9 hours' traffic video data obtained from the Louisiana Department of Transportation and Development. Experimental results demonstrate that the combination of CenterNet and Deep SORT, Detectron2 and Deep SORT, and YOLOv4 and Deep SORT produced the best overall counting percentage for all vehicles.

*Index Terms*—Deep learning, Object Detection, Tracking, Vehicle Counts

## I. INTRODUCTION

Accurate estimation of the number of vehicles on the road is an important endeavor in intelligent transportation system (ITS). An effective measure of on-road vehicles can have a plethora of application in transportation sciences including traffic management, signal control and on-street parking [2, 13, 11]. Technically, most vehicle counting methods are characterized into either hardware or software-based systems [14]. Inductive-loop detectors and piezoelectric sensors are the two most extensively used hardware systems till date. Although they have higher accuracies than software based systems, they are intrusive and expensive to maintain. On the other hand, software based system that use video cameras and run on computer vision algorithms present an inexpensive and non-intrusive approach to obtain vehicle counts. Similarly, with increasing computing capabilities and recent successes in object detection and tracking technology, they manifest a tremendous potential to surrogate hardware based systems. Part of the reason to make such a claim is due to the rapid advancement in the field of deep learning and high performance computing, which has fueled an era of ITS within the multi-disciplinary arena of transportation sciences.

This study is motivated by the need to present a robust vision-based counting system that addresses the challenging real-world vehicle counting problem. The visual understanding of objects in an image sequence must face many challenges, perhaps customary to every counting task such as difference in scales and perspectives, occlusions, illumination effects and many more [7]. To address these challenges, several deep learning based techniques are proposed to accurately detect and count the number of vehicles in different environmental conditions. Out of all the problems associated to counting, one that stands out the most would be the occlusion in traffic videos. They appear quite frequently on most urban roads that experience some form of congestion. This leads to ambiguity in vehicle counting which could likely undermine the quality of traffic studies that rely on vision-based counting schemes to estimate traffic flows or volumes [19]. One of the objectives of this paper is to propose a counting system that is robust to occlusion problem and can provide a resolve in accurately counting vehicles that experience multi-vehicle occlusion.

Passenger cars occupy the greatest proportion of on-road vehicles and most often than not they get occluded by trucks when they are either too near or distant to traffic cameras. Therefore, the scope of this study is limited to counting cars and trucks only. We focus on real-time vehicle tracking and counting using state-of-the-art object detection and tracking algorithms. The rest of the paper is outlined as follows: Section 2 briefly reviews related works in the field of vehicle counting. Section 3 contains data description. Section 4 describes the proposed methodology including different object detection and tracking algorithms. Section 5 includes empirical results, and Section 6 details the conclusions of this study.

## II. RELATED WORK

Vision-based vehicle counting is an interesting computer vision problem tackled by different techniques. As per the taxonomy accepted in [26], the counting approach could be broadly classified into five main categories: counting by frame-differencing [24, 8], counting by detection [29, 23], motion

Vishal Mandal is with the Department of Civil and Environmental Engineering, University of Missouri-Columbia and with WSP USA, 211 N Broadway Suite # 2800, St. Louis, MO 63102 USA (e-mail: vmghv@mail.missouri.edu).

Yaw Adu-Gyamfi is with the Department of Civil and Environmental Engineering, E2509 Lafferre Hall, Columbia, MO 65211 USA (e-mail: adugyamfiy@missouri.edu).

based counting [22, 5, 6, 16], counting by density estimation [12] and deep learning based counting [25, 27, 18, 28, 1, 21, 10]. The first three counting methods are environmental sensitive and generally don't perform very well in occluded environments or videos with low frame rates. While counting by density estimation follows a supervised approach, they perform poorly in videos that have larger perspective and contain oversized vehicles. Density estimation based methods are also limited in their scope of detection and lack object tracking capabilities. Finally, out of all these counting approaches, deep learning based counting techniques have had the greatest developments in recent years. The advancement in their built architectures have significantly improved the vehicle counting performance. In this study, we mainly focus on studying counting methods that are founded on deep learning based architectures.

Awang et al. in [3] proposed a deep learning based technique that tabulates the number of vehicles based on the layer skipping-strategy within a convolutional neural network (CNN). Prior to performing counts, their approach classifies the vehicle into different classes based on their distinct features. Dai et al. in [9] deployed a video based vehicle counting technique using a three-step strategy of object detection, tracking and trajectory analysis. Their method uses a trajectory counting algorithm that accurately computes the number of vehicles in their respective categories and tracks vehicle routes to obtain traffic flow information. Similarly, a deep neural network is trained to detect and count the number of cars in [17]. This approach integrates residual learning alongside inception-style layers to count cars in a single look. Lately, it has been demonstrated that single-look techniques have the potential to excel at both speed and accuracy [20] requirements useful for object recognition and localization. This could also, prove beneficial to process image frames at much faster rates that can accurately produce vehicle counts in real-time conditions. The authors in [15] deliberate counting as a computer vision problem and present an adaptive real-time vehicle counting algorithm that takes robust detection and counting approach in an urban setting.

Although video-based counting systems have emerged as an active research area, there are issues with detection and re-identification of vehicles while they cross each other in separate road lanes. To counter this problem, Bui et al. in [4] successfully deployed state-of-the-art YOLO and SORT algorithms to perform vehicle detection and tracking respectively. To further improve their video-based vehicle counter, they followed a distinguished region tracking paradigm that works well for intricate vehicle movement scenarios. Generally, most object counting literature [27, 18, 28, 1] approximates the object densities, maps them and computes densities over the entire image space to obtain vehicle counts. However, the accuracy of these methods drop whenever a video has a larger perspective or if a large bus or truck appears. The FCN-rLSTM network proposed in [26] tackles problems associated to larger perspective videos by approximating vehicle density maps and performing vehicle counts by integrating fully convolutional neural networks (FCN) with long short term memory networks (LSTM) in a residual learning environment. This approach leverages the capabilities of FCN based pixel wise estimation and the strengths of LSTM to learn difficult time-based vehicle dynamics. The counting accuracy is thus, improved by putting the time-based correlation into perspective.

### III. DATA DESCRIPTION

Traffic images and video feeds were the two kinds of dataset used in this study. These datasets were obtained from the cameras located at 6 different roadways over a seven-day period. The cameras were installed across different roadways in New Orleans and maintained by the Louisiana Department of Transportation and Development (La DOTD). To train and generate robust models, datasets pertaining to different weather conditions were collected. To incorporate that, video feeds were recorded at the start of every hour for one minute and followed the same loop for the entire 24 hours in a day. This recording was further continued for 1 week at all the 6 roadways. Traffic images and videos consist of daylight, nighttime and rain. To train all the models used in this study, altogether 11,101 images were manually annotated for different classes of vehicles viz. cars and trucks. Figure 1 shows all the 6 different cameras maintained by La DOTD and their respective camera views. Similarly, any vehicle that travelled across those green and blue polygons were counted and appended in the north and southbound directions respectively.

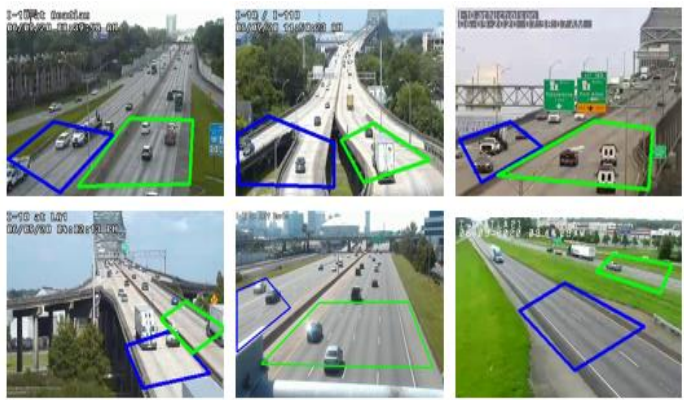

Fig. 1. Camera Locations

### IV. METHODOLOGY

The study compares the combination of different object detectors and trackers for performing vehicle counts. As seen from Figure 2, the proposed vehicle counting framework initiates by manually annotating traffic images. This is followed by training several object detection models which can then be used to detect different classes of vehicles. All the object detection models are trained on NVIDIA GTX 1080Ti GPU.

After obtaining detection results for each video frame, different tracking algorithms are used for multi-object tracking. In this study, we used both online and offline tracking algorithms. Although offline tracking algorithms yield better results, the advantage of using online trackers could be realized in applications that involve online traffic control scenarios. Similarly, based on the detection outcomes, each vehicle is counted only once as per their trajectory matching function' intrinsic to every object tracking algorithm. The green and blue polygons drawn on the cameras (see Fig. 1) assigns the entrance and exit zones for every vehicles' trajectory and computes the number of vehicles passing through the north and southbound directions respectively. Altogether, 4 different state-of-the-art object detectors and trackers were used making a total of 16 different detector-tracker combinations. Upon obtaining vehicle counts, all these detector-tracker combinations were further analyzed and had their performance capabilities compared based off different environmental conditions. The object detectors and tracking algorithms used in this study are further explained in detail in the subsequent sections.

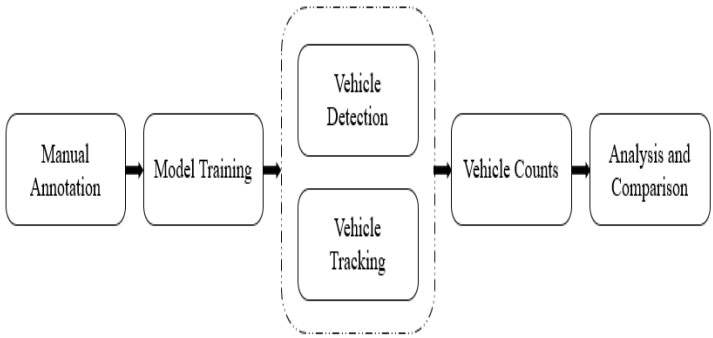

Fig. 2. Detection-Tracking based Vehicle Counting Framework

## A. OBJECT DETECTORS

1. CenterNet

With the advancement in deep learning, object detection algorithms have significantly improved. In this study, the authors implemented an object detection framework called CenterNet [30] which discovers visual patterns within each section of a cropped image at lower computational costs. Instead of detecting objects as a pair of key points, CenterNet detects them as a triplet thereby, increasing both precision and recall values. The framework builds up on the drawbacks encountered by CornerNet [31] which uses a pair of corner-keypoints to perform object detection. However, CornerNet fails at constructing a more global outlook of an object, which CenterNet does by having an additional keypoint to obtain a more central information of an image. CenterNet functions on the intuition that if a detected bounding box has a higher Intersection over Union (IoU) with the ground-truth box, then the likelihoods of that central keypoint to be in its central region and be labelled in the same class is high. Hence, the knowledge of having a triplet instead of a pair increases CenterNet's superiority over CornerNet or any other anchor-based detection approaches. Despite using a triplet, CenterNet is still a single-stage detector but partly receives the functionalities of RoI pooling. Figure 3 shows the architecture of CenterNet where it uses a CNN backbone that performs cascade corner pooling and center pooling to yield two corner and a center keypoint heatmap. Here, cascade corner pooling enables the original corner pooling module to receive internal information whereas center pooling helps center keypoints to attain further identifiable visual pattern within objects that would enable it to perceive the central part of the region. Likewise, analogous to CornerNet, a pair of detected corners and familiar embeddings are used to predict a bounding box. Then after, the final bounding boxes are determined using the detected center keypoints. In this study, CenterNet was trained on NVIDIA GTX 1080Ti GPU which took approximately 22 hours.

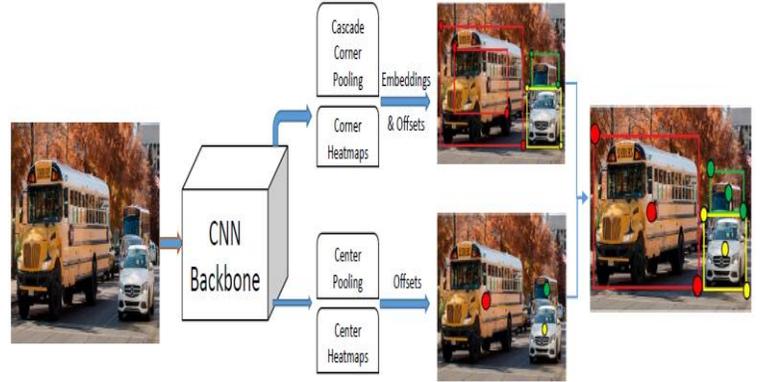

Fig. 3. Architecture of CenterNet

2. Detectron2

Detectron2 [32], a deep neural network builds up on the Mask R-CNN benchmark, capable of implementing state-of-the-art object detection algorithms. Fueled by the PyTorch deep learning framework, it includes features such as panoptic segmentation, dense-pose, Cascade R-CNN, rotated bounding boxes, etc. To perform object detection and segmentation, Detectron2 requires images and its annotated database to follow annotation format as followed by the COCO dataset. The annotation consists of every individual object present in all images of the training database. Detectron2 supports implementation to multiple object detection algorithms using different backbone network architectures such as ResNET {50, 101, 152}, FPN, VGG16, etc. Hence, it can be used as a library to support a multitude of projects on top of it. In this study, all vehicles including cars and trucks present in the image is hand-annotated for higher training precision. Similarly, while performing vehicle detection, Detectron2 uses focused detection step comprising of scanning the regions of interest in a pixel-wise manner and performing prediction with the help of a mask. An advantage of using Detectron2 is that it learns and trains at a much faster rate. Detectron2 shared the same hardware resources like CenterNet and took approximately 36 hours to train.

3. YOLOv4

You Only Look Once (YOLO) is the state-of-the-art object detection algorithm. Unlike traditional object detection systems, YOLO investigates the image only once and detects if there are any objects in it. Out of all the earlier versions of YOLO, YOLOv4 is the latest and most advanced iteration till date [33]. It has the fastest operating speed for use in production systems and for optimization in parallel computations. Some of the new techniques adopted in YOLOv4 are: (i) Weighted-Residual-Connections, (ii) Cross-Stage-Partial-Connections, (iii) Cross mini-batch, (iv) Normalization (CmBN), (v) Self-adversial-training, (vi) Mish-activation, etc. To obtain higher values for precision, YOLOv4 uses a Dense Block, a deeper and more complex network. Similarly, the backbone of its feature extractor uses CSPDarknet-53, which deploys the CSP connections alongside Darknet-53 from the earlier YOLOv3. In addition to CSPDarknet-53, the architecture of YOLOv4 comprises of SPP additional module, PANet path-aggregation neck and YOLOv3 anchor-based head. The SPP block is stacked over CSPDarknet53 to increase the receptive field that could discretize the most remarkable context features and makes sure that there is no drop in its network operation speed. Similarly, PANet is used for parameter aggregation from several levels of backbone in place of Feature Pyramid Network (FPN) that is used in YOLOv3. YOLOv4 models took approximately 24 hours to train and shared the same hardware resources with CenterNet and Detectron2.

4. EfficientDet

EfficientDet is a state-of-the-art object detection algorithm that basically follows single-stage detectors pattern [34]. The architecture of EfficientDet is shown in Figure 4. Here, the ImageNet-pretrained EfficientNets has been deployed as the network's backbone. Similarly, in order to obtain an easier and quicker multi-scale fusion of features, a weighted bi-directional feature pyramid network (BiFPN) has been proposed. BiFPN here, serves as the feature network and receives approximately 3-7 features from the backbone network and continually performs top-down and bottom-up bidirectional fusion of features. These synthesized features are transferred to a class and box network to achieve vehicle class and bounding box predictions correspondingly. Also, all the vehicle class and box network weights are jointly shared across every feature level. Similarly, to achieve higher accuracy, a new compound scaling method is proposed for EfficientDet. This compound scaling approach scales up the overall dimensions of width, depth, backbone resolution, BiFPN along with box and class prediction networks. Although, the primary goal of EfficientDet was to perform object detection, it could also be deployed to perform tasks such as semantic segmentation. Training an EfficientDet model took approximately 36 hours on an NVIDIA GTX 1080Ti GPU.

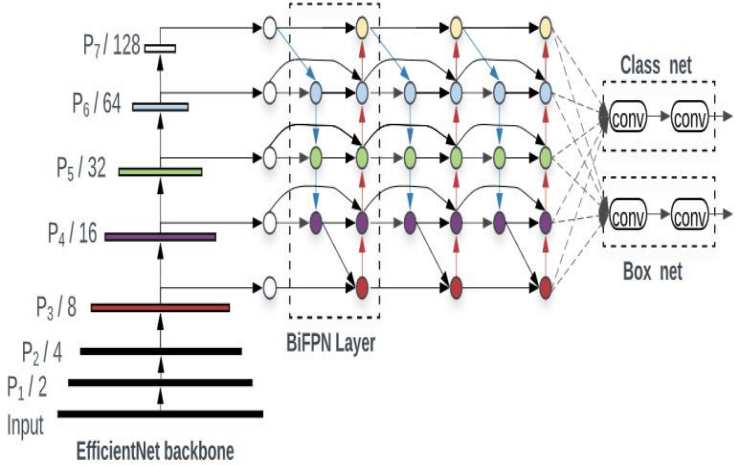

Fig. 4. Architecture of EfficientDet [34]

## B. OBJECT TRACKER

### 1. IOU Tracker

IOU tracker is built on the assumption that every object is tracked on a per-frame basis such that there are none or very few gaps present in between detections [35]. Similarly, IOU assumes that there is a greater overlap value for intersection over union while obtaining object detections in successive frames. The equation (1) measures the Intersection over Union which forms the basis for this approach.

---

Algorithm 1 IOU Tracking

---

1: function Tracker( detections, $\sigma_l$, $\sigma_h$, $\sigma_{iou}$, $min_{tsize}$
◁ detection(dict(class, score, box))
2:   let $\sigma_l$ ← low detection threshold
3:   let $\sigma_h$ ← high detection threshold
4:   let $\sigma_{iou}$ ← IOU threshold
5:   let $min_{tsize}$ ← minimum track size in frames
6:   let $T_a$ ←[]                                      ◁ active tracks
7:   let $T_f$ ←[]
◁ finished tracks
8:   for $frame, dets$ in detections do
9:     dets ← filter for dets with score ≥ $\sigma_l$
10:    let $T_u$ ←[]                                    ◁ updated tracks
11:    for $t_i$ in $T_a$ do
12:    if not empty(dets) then
13:      $b_{iou}$, $b_{box}$ ← find max iou box(tail box($t_i$), dets)
14:      if $b_{iou}$ ≥ $\sigma_{iou}$  then
15:    append new detection($t_i$, $b_{box}$)
16:    set max score($t_i$, box score($b_{box}$))
17:    set class($t_i$, box class($b_{box}$))
18:      $T_u$ ← append($T_u$, $t_i$)
19:      remove(dets, $b_{box}$)                        ◁ remove box from dets
20:    if empty($T_u$) or $t_i$ is not last($T_u$) then
21:      if get max score($t_i$)≥ $\sigma_h$ or size($t_i$)≥ $min_{tsize}$ then
22:      $T_f$ ← append($T_f$, $t_i$)
23:    $T_n$ ← new tracks from dets
24:    $T_a$ ← $T_u$ + $T_n$
25:      return $T_f$

---

$$IOU\ (a,b) = \frac{Area\ (a) \cap Area\ (b)}{Area\ (a) \cup Area\ (b)} \qquad (1)$$

IOU tracker specifically tracks objects by assigning detection with the highest IOU value (equation 1) to the last detection in the earlier frame if a specific threshold value is satisfied. In cases, where any detection was not assigned to an existing track, then it begins with a new one. Likewise, any track that was devoid of an assigned detection will end. Since, we aim to track vehicles in this study, the IOU performance could be further enhanced by canceling tracks that don't meet a certain threshold time length and where no detected vehicle exceeded the required IOU threshold. It is important to note that IOU tracker is heavily reliant on how accurately vehicles are recognized by object detection models, so special focus should be laid out on effectively training object detection algorithms. IOU's ability to handle frame rates of over 50,000 fps in conjunction to its low computational cost makes it an incredibly powerful object tracker. The step-wise operations followed by IOU tracker is shown in Algorithm 1.

Similarly, Kalman-IOU (KIOU) tracking has been further explored. The Kalman filter's ability of performing predictions allows users to skip frames while still keeping track of the object. Skipping frames allows the detector to speed-up the process as in a tracking-by-detection task, smaller number of frames wedges lower computational requirement. Using an appropriate object detector with Kalman-IOU tracker, and configuring the frames to skip two-thirds of frames per second could enable the tracker to run in real-time. Likewise, this feature could also improve the performance of Kalman-IOU tracker compared to the original IOU tracker.

2. SORT

Simple Online and Realtime Tracking (SORT) is an implementation of tracking-by-detection framework where the main objective is to detect objects each frame and associate them for online and real-time tracking application [36]. Methods such as Kalman Filter and Hungarian algorithm are used for tracking. The characteristic feature of SORT is that it only uses detection information from the previous and current frames, enabling it to competently perform online and real-time tracking. To further explain this, an object model is described as expressed in equation (2) where *u*, *v*, *s*, and *r* represent the horizontal pixel location, vertical pixel location, area and aspect ratio of the target object respectively. Anytime a detection is linked to a target object, the detected bounding box is used to inform the target state and the horizontal and velocity values are solved using Kalman filters. This helps in identifying target's identity in successive frames and facilitates tracking.

$$X = [\ u,\ v,\ s,\ r,\ \dot{u},\ \dot{v},\ \dot{s}\ ]^T \quad (2)$$

3. Feature Based Object Tracker

In Feature-based object tracking, there is the usage of appearance information to track objects in respective traffic scenes. This method is useful in tracking vehicles in occluded settings. The system extracts object features from one frame and then matches appearance information with successive frames based on the measure of similarity. Feature-based object tracking consists of both feature extraction and feature correspondence. The feature points are extracted from the objects in an image using various statistical approaches. Feature correspondence is considered an arduous task since, a feature point in one image may have analogous points in other images which could perhaps, pose ambiguity problems in feature correspondence. To undermine ambiguity, most contemporary algorithms use exhaustive search along with correlation over larger pixels of image neighborhood. Likewise, the minimum value of cosine distance is also useful at computing any resemblance between some of the characteristic features which is useful for object tracking. In the current study, a feature-based object tracker called Deep SORT is deployed. Some of the features of this tracking algorithm is explained in detail as follows.

3.1 Deep SORT

The Simple Online and Realtime Tracking with a Deep Association metric (Deep SORT) enables multiple object tracking by integrating appearance information with its tracking components [37]. A combination of Kalman Filter and Hungarian algorithm is used for tracking. Here, Kalman filtering is performed in image space while Hungarian technique facilitates frame-by-frame data association using an association metric that computes bounding box overlap. To obtain motion and appearance information, a trained convolutional neural network (CNN) is applied.

By integrating CNN, the tracker achieves greater robustness against object misses and occlusions while preserving the trackers ability to quickly implement to online and realtime scenarios intact. The CNN architecture of the system is shown in Table I. A wide residual network with two convolutional layers followed by six residual blocks is applied. In dense layer 10, a global feature map of dimensionality 128 is calculated. Finally, batch and $\ell_2$ normalization features over the unit hypersphere accesses compatibility with cosine arrival metric. Overall, Deep SORT is a highly versatile tracker and can match performance capabilities with other state-of-the-art tracking algorithms.

TABLE I.
OVERVIEW OF DEEP SORT'S CNN ARCHITECTURE

| Name | Patch Size/Stride | Output Size |
|---|---|---|
| Conv1 | 3 × 3/1 | 32 × 128 × 64 |
| Conv2 | 3 × 3/1 | 32 × 128 × 64 |
| Max Pool 3 | 3 × 3/2 | 32 × 64 × 32 |
| Residual 4 | 3 × 3/1 | 32 × 64 × 32 |
| Residual 5 | 3 × 3/1 | 32 × 64 × 32 |
| Residual 6 | 3 × 3/2 | 64 × 32 × 16 |
| Residual 7 | 3 × 3/1 | 64 × 32 × 16 |
| Residual 8 | 3 × 3/2 | 128 × 16 × 8 |
| Residual 9 | 3 × 3/1 | 128 × 16 × 8 |
| Dense 10 | | 128 |
| Batch and $\ell_2$ normalization | | 128 |

## V. RESULTS

This section evaluates the performance of different combinations of object detectors and trackers. The main goal of this study is to identify the best performing object detector-tracker combination. For comparative analysis, the models are tested on a total of 546 video clips of length 1 minute each comprising of over 9 hours' total video length. Figure 1 shows all the camera views with manually generated green and blue polygons that record the number of vehicles passing through them in both north and southbound directions respectively. The vehicle counts are evaluated based on four different categories: (i) overall count of all vehicles, (ii) total count of cars only, (iii) total count of trucks only, and (iv) overall vehicle counts for different times of the day (i.e. daylight, nighttime, rain). To establish ground truth, all the vehicles are manually counted from the existing 9 hours' video test data. The performance is assessed by expressing the automatic counts obtained from different model combinations over the ground truth value expressed in per hundredth or percentage.

To examine the performance of object detectors, heat maps showing False Negatives (FN), False Positives (FP) and True Positives (TP) are plotted in Figure 5 for all the object detectors used in the study. The models were tested on altogether 6 camera views at different times of the day. The top left and right columns show the heat maps generated for CenterNet and Detectron2 while the bottom left and right columns display heat maps produced for EfficientDet and YOLOv4 respectively. For all these respective object detectors, the first column represents FN, the second column designates FP and the third column denotes TP. The detection is classified as False Negative (FN) if the detector fails at detecting the vehicle despite it being present at that spot. Therefore, the column showing FN should necessarily not have brighter intensity of colors around those sections of the roadway. Almost all object detectors have performed well at detecting FN in most camera views except for a few instances where Detectron2 in its last camera row recorded sharp intensity of heat scales in its south bound direction and for CenterNet's 5$^{th}$ camera view where it generates heat maps in its south bounds as well. This is largely because certain camera views had insufficient number of traffic images used for training and could have possibly experienced heavy congestion at those sites. For instance, heat maps closer to the camera in night views are produced generally when the heavy gross vehicles such as buses and trucks remain congested at those spots for a very long time.

Similarly, the detection is classified as False Positive (FP) if the detector erroneously detects a vehicle at a spot with no vehicles present. As observed from the heat maps, the FP columns for object detectors are generally clean with a few camera views in Detectron2 and EfficientDet generating incorrect classifications. The camera view with flyovers or overpass roads caused the model to misclassify some of the detections. Sometimes, camera movements and conditions such as rain sticking on the camera lens and pitch darkness also cause such misclassifications. Ideally, we do not aim at seeing intense heat maps for both false negatives and false positives. However, if we have higher false positives but obtain lower false negatives, then the model might have been too confident which is not very ideal. Finally, True Positive (TP) is the one that correctly detects vehicle when there are any actual vehicles present on the roadways. Most object detection models generated correct true positives except for a few camera views where the vehicles are either too distant or encounters lowlight or nighttime conditions where only the vehicles' headlights were visible.

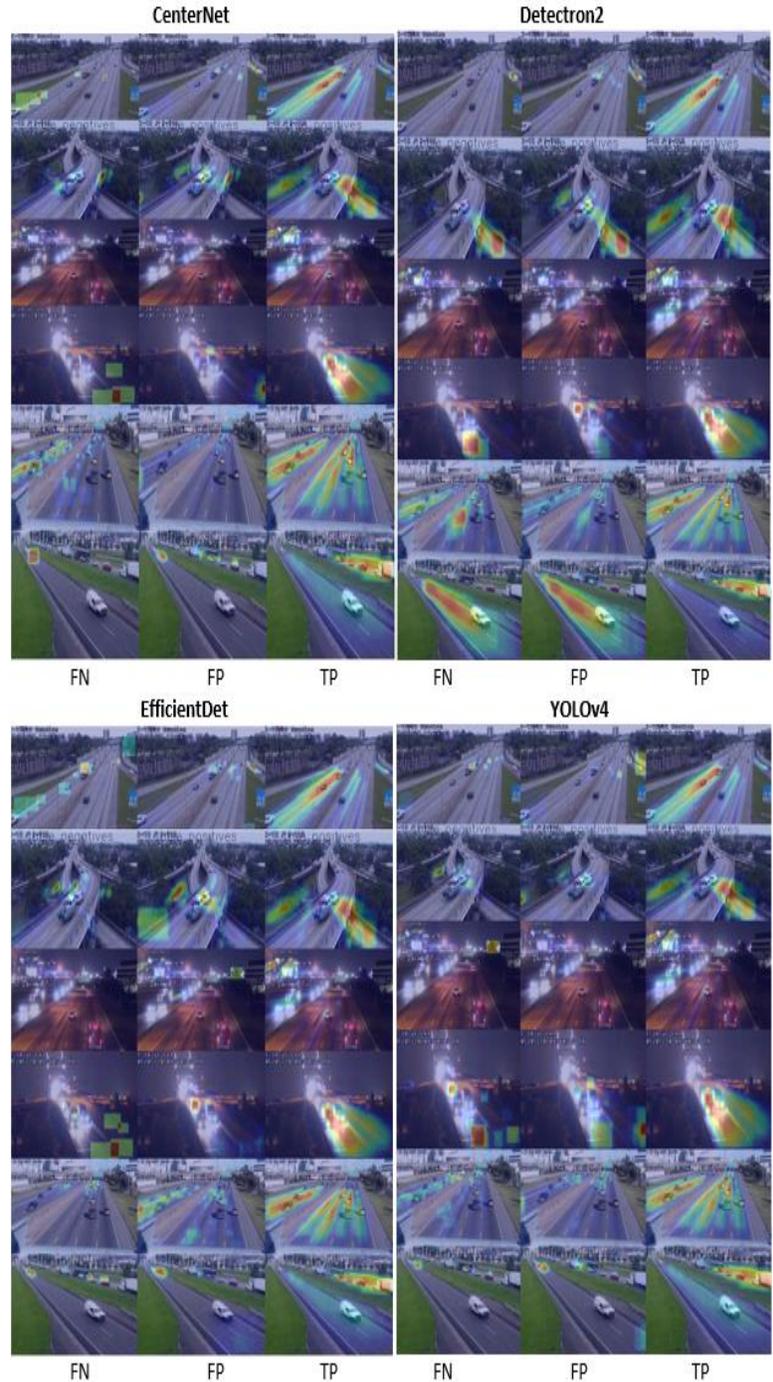

Fig. 5. Heat Maps Generated for Different Object Detectors

Figure 6 shows the overall count percentage for all vehicles. As seen from the figure, the overall count percentage for some of the model combinations exceed over 100 percent while a couple combinations obtain counting results below 45 percent of the actual counts. Any model combination that either over-counts or under-counts the actual number of vehicles are considered faulty match while the ones that perform counts in the order closer to 100 percent are termed an optimal match. The best performing model combinations that obtained a more accurate count estimate for all vehicles were YOLOv4 and Deep SORT, Detectron2 and Deep SORT, and CenterNet and Deep SORT. Thus, all these model combinations can be considered an optimal match.

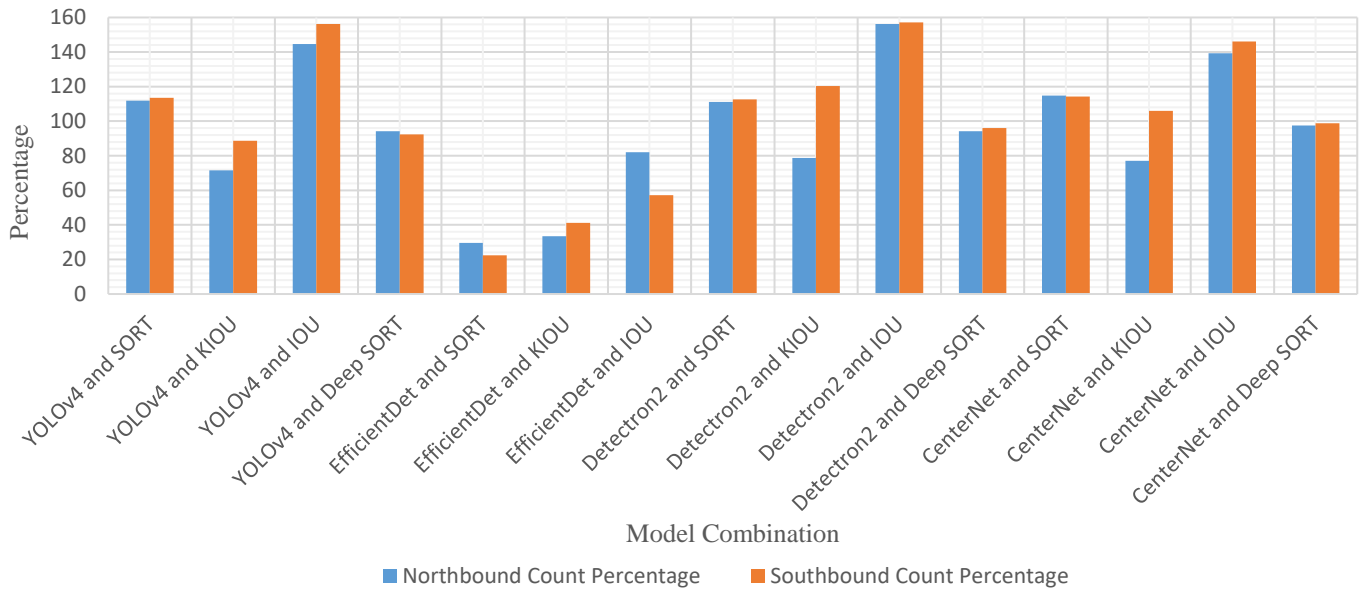

Fig. 6. Performance of Model Combination for All Vehicles Count

Similarly, Figures 7-8 compares the performance of different model combinations for counting cars and trucks respectively. From Figure 7, it can be observed that CenterNet and IOU, CenterNet and SORT, Detectron2 and SORT, EfficientDet and Deep SORT, and YOLOv4 and Deep SORT obtained the best counting results. These detector-tracker combinations performed well in both north and southbound directions respectively. Occlusion issues created a hindrance in correctly locating cars which would often be obstructed by larger vehicles whenever they are too close to the camera. Likewise, in Figure 8, the truck counter performance is assessed. Out of all the model combinations, only EfficientDet and Deep SORT obtained acceptable counting performance in both north and southbound directions. Although, the combination of CenterNet and KIOU, and EfficientDet and SORT separately obtained accurate counting results, their scope was limited to only either North or Southbound directions respectively. Most of the other model combinations didn't accurately count trucks due to the presence of other heavy gross vehicles (HGV) such as buses, trailers, and multi-axle single units. These HGVs often confused the models and were assigned as trucks that generated an over-estimate of truck counts. Exaggerating the actual number of vehicle counts (either trucks or cars) could be attributed to that fact that some of the detectors produced multiple bounding boxes for the same vehicle while traversing the video scene. This impelled the tracker to confuse the same vehicle as different ones and assign them with newer values every time a bounding box re-appears.

Likewise, Table II illustrates the performance comparison of models in different weather conditions. The counting results show that the best performing model combinations were YOLOv4 and Deep SORT, Detectron2 and Deep SORT, and CenterNet and Deep SORT, analogous to the comparison chart as shown in Figure 6. Vehicle counting accuracies largely depends on the precision of object detection models. However, it is evident from Table II that the models didn't achieve optimal results for the most part. The reasons could be partly attributed to the inferior camera quality, unstable camera views due to the wind blowing on highways, and the presence of fog or mist on camera lens.

During daylight, nighttime and rainy conditions, EfficientDet's combination with SORT and KIOU failed miserably at counting the number of vehicles. EfficientDet mainly suffered with its detection capability. For model combinations that recorded count percentage over 100 typically had both detector and tracker at fault. Object detectors generated multiple bounding boxes for the same vehicle that resulted in over-counting of the number of vehicles. Also, some of the trackers did not perform ideally at predicting vehicle trajectories and assigned them as separate vehicles at certain occasions.

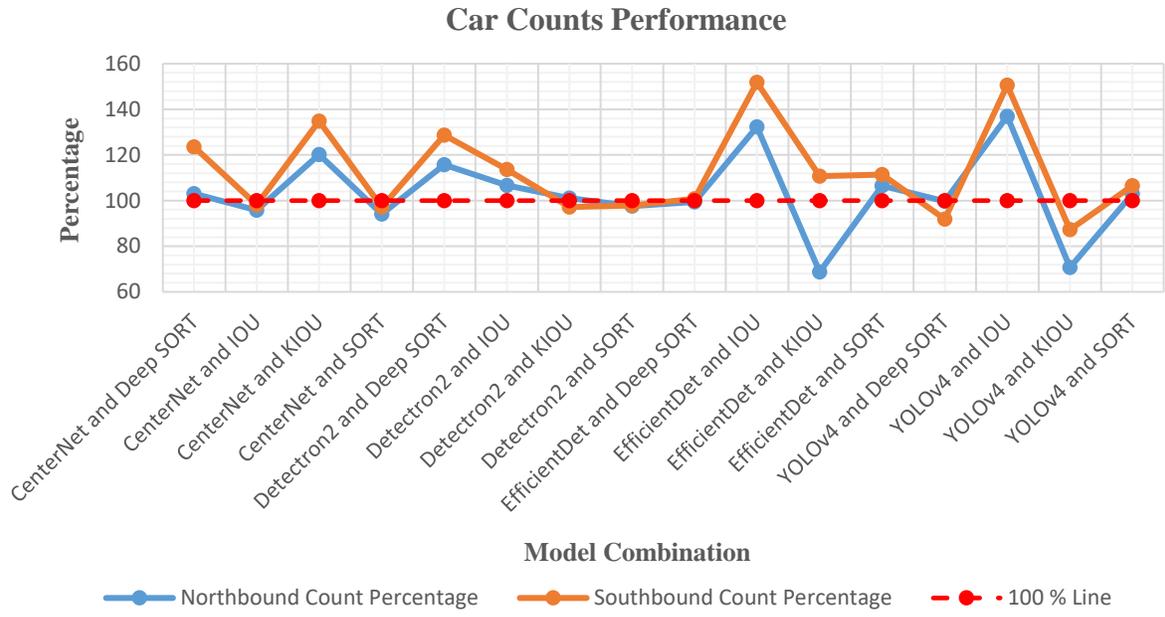

Fig. 7. Performance of Model Combination for Car Counts Only

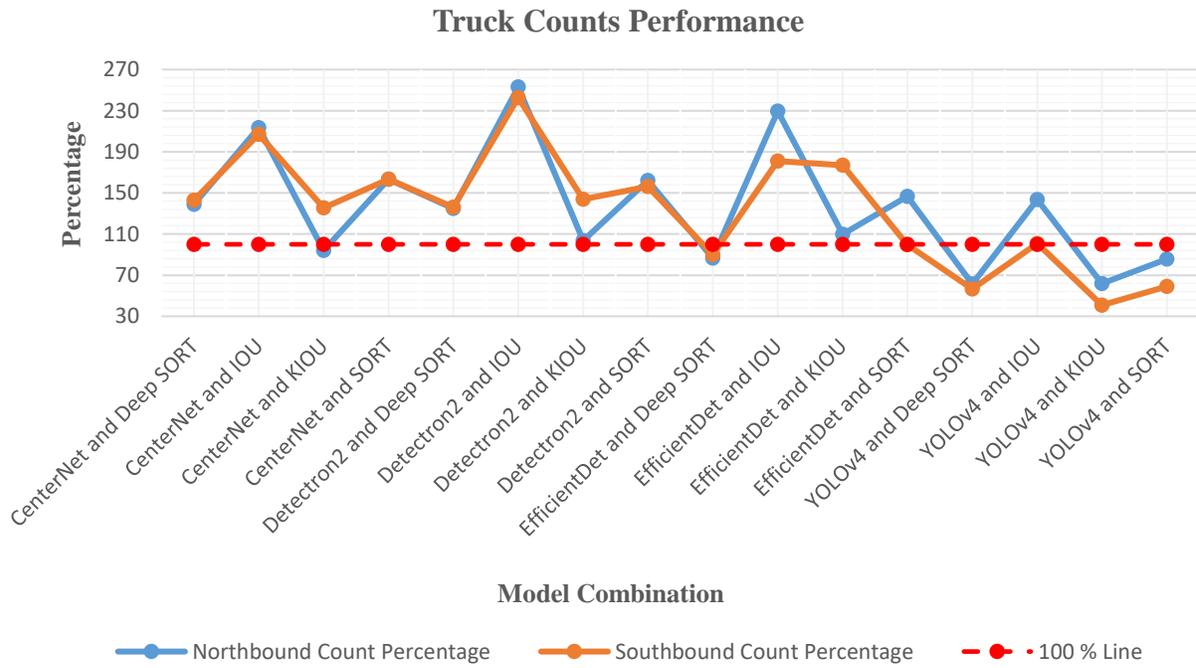

Fig. 8. Performance of Model Combination for Truck Counts Only

TABLE II.
PERFORMANCE OF MODEL COMBINATIONS IN DIFFERENT WEATHER CONDITIONS

| Time of Day | Model Combination | Northbound Count Percentage | Southbound Count Percentage |
|---|---|---|---|
| | YOLOv4 and SORT | 112.3597165 | 114.9770576 |
| | YOLOv4 and KIOU | 70.81364442 | 89.70461715 |
| | YOLOv4 and IOU | 144.3812758 | 155.2767422 |
| | YOLOv4 and Deep SORT | 92.277562 | 91.5865623 |
| | EfficientDet and SORT | 30.53610906 | 23.24962286 |
| | EfficientDet and KIOU | 32.47185667 | 41.27978954 |
| | EfficientDet and IOU | 82.04832193 | 57.673148 |
| | Detectron2 and SORT | 110.6098641 | 114.0952108 |
| | Detectron2 and KIOU | 76.68340224 | 121.243189 |
| | Detectron2 and IOU | 153.7507383 | 153.6062518 |
| | Detectron2 and Deep SORT | 94.30005907 | 97.24691712 |
| | CenterNet and SORT | 114.2941524 | 115.5147691 |
| | CenterNet and KIOU | 75.02215003 | 105.6638945 |
| | CenterNet and IOU | 137.0496161 | 144.063665 |
| Daylight | CenterNet and Deep SORT | 97.42321323 | 100.1362202 |
| | YOLOv4 and SORT | 107.1243523 | 106.5976714 |
| | YOLOv4 and KIOU | 72.99222798 | 87.12807245 |
| | YOLOv4 and IOU | 145.9196891 | 166.2354463 |
| | YOLOv4 and Deep IOU | 91.256962 | 90.25664125 |
| | EfficientDet and SORT | 59.45595855 | 55.62742561 |
| | EfficientDet and KIOU | 36.1952862 | 36.01368691 |
| | EfficientDet and IOU | 76.52011225 | 53.14617619 |
| | Detectron2 and SORT | 110.5569948 | 106.2742561 |
| | Detectron2 and KIOU | 82.83678756 | 117.076326 |
| | Detectron2 and IOU | 166.9689119 | 184.152652 |
| | Detectron2 and Deep SORT | 93.84715026 | 93.5316947 |
| | CenterNet and SORT | 110.880829 | 107.7619664 |
| | CenterNet and KIOU | 74.74093264 | 112.4191462 |
| | CenterNet and IOU | 144.753886 | 161.3842173 |
| Night-time | CenterNet and Deep SORT | 95.98445596 | 92.94954722 |
| | YOLOv4 and SORT | 114.4578313 | 101.9874477 |
| | YOLOv4 and KIOU | 82.06157965 | 74.89539749 |
| | YOLOv4 and IOU | 145.9170013 | 153.7656904 |
| | YOLOv4 and Deep SORT | 91.258975 | 89.256987 |
| | EfficientDet and SORT | 46.18473896 | 47.90794979 |
| | EfficientDet and KIOU | 49.45567652 | 46.2195122 |
| | EfficientDet and IOU | 92.32 | 55.47169811 |
| | Detectron2 and SORT | 121.9544846 | 101.5690377 |
| | Detectron2 and KIOU | 108.4337349 | 112.2384937 |
| | Detectron2 and IOU | 177.5100402 | 165.0627615 |
| | Detectron2 and Deep SORT | 94.77911647 | 81.48535565 |
| | CenterNet and SORT | 131.8607764 | 107.1129707 |
| | CenterNet and KIOU | 119.1432396 | 99.47698745 |
| | CenterNet and IOU | 169.7456493 | 150.3138075 |
| Rain | CenterNet and Deep SORT | 102.0080321 | 87.23849372 |

## VI. CONCLUSION

In this study, a detection-tracking framework is applied to automatically count the number of vehicles on roadways. The state-of-the-art detector-tracker model combinations have been further refined to achieve significant improvements in vehicle counting results although there are still many shortcomings which the authors aim to address in the future study. Occlusion and lower visibility created identity switches and same vehicles were detected multiple times which caused the model to sometimes over-exaggerate the number of vehicles. Although, conditions such as inferior camera quality, occlusion and low light conditions proved tricky in accurately detecting different classes of vehicles, certain combinations of detector-tracker framework functioned fine for challenging conditions as well. Deep learning based object detection models coupled with both online and offline multi-object tracking systems could integrate real-time object detections in conjunction to tracking vehicle movement trajectories. This outline was accepted which in turn facilitated accurate vehicle counts. Moreover, we experimented with the detector-tracker ability to correctly detect different classes of vehicles, estimate vehicles' speed, direction and its trajectory information to identify some of the best performing models which could be further fine-tuned to remain robust at counting vehicles in different directions and environmental conditions. The figures and tables present a systematic representation of what model combinations perform well at obtaining vehicle counts in different conditions. Overall, for counting all vehicles on the roadway, experimental results from this study prove that YOLOv4 and Deep SORT, Detectron2 and Deep SORT, and CenterNet and Deep SORT were the most ideal combinations.